# GREEN ECONOMIC LOAD DISPATCH: A REVIEW AND IMPLEMENTATION

**Shahbaz Hussain**
University of Engineering and Technology, Lahore 54890, Pakistan
shahbazpec@yahoo.com

***ABSTRACT:*** *The economic dispatch of generators is a major concern in thermal power plants that governs the share of each generating unit with an objective of minimizing fuel cost by fulfilling load demand. This problem is not as simple as it looks because of system constraints that cannot be neglected practically. Moreover, increased awareness of clean technology imposes another important limit on the emission of pollutants obtained from burning of fossil fuels.*
*Classical optimization methods lack the ability of solving such a complex and multi-objective problem. Hence, various modern artificial intelligence (AI) techniques based on evolution and social behaviour of organisms are being used to solve such problems because they are easier to implement, give accurate results and take less computational time.*
*In this work, a study is done on most of the contemporary basic AI techniques being used in literature for power systems in general and combined economic emission dispatch (CEED) in particular. The dispatch problem is implemented on IEEE 30-bus benchmarked system in MATLAB for different load demands considering all gases ($CO_X$, $NO_X$ and $SO_X$) using particle swarm optimization (PSO) and genetic algorithm (GA) and their results are compared with each other.*

## 1. INTRODUCTION

The goal of generators dispatch in a thermal power plant is to deliver at optimal point by satisfying system constraints for economy saving. The total generation should meet the load demand of consumers and also the line losses if they are considered. To achieve this objective, generators fuel cost which constitutes mainly the generation cost is required and this data is achieved from heat rate curves of generators. Further, heat rate curves are converted into cost rate curves and these curves can be mathematically approximated to quadratic equations in terms of generators real power for the purpose of programming [1].

The system constraints, mainly focused are generator limits and power balance. Each generator has its minimum and maximum limits for producing power beyond which it cannot deliver and operation out of these limits is highly dangerous and uneconomical for the system. Hence, it is always ensured to keep the generation of each generator within allowable range. The second constraint is that the generated power should match the load demand, and the losses happening in the transmission lines if considered.

Environmental regulatory authorities have been very active in the past few decades and impose strict laws concerning pollution emitted by industrial sector. Hence, now the objective is not only to save the fuel cost but another objective of pollutants emission within said limits comes into picture. This makes it a complex multi objective problem as both objectives of optimizing economy and emission are conflicting in nature. $CO_X$, $NO_X$ and $SO_X$ are the main gases that pollute the environment during combustion of fossil fuels in thermal power plants. The emission characteristics of generators can also be mathematically represented as quadratic equations in terms of generators real power [1].

Combined economic emission dispatch puts together the optimization of fuel cost and emission. These two objectives can be made a single objective function by imposing penalty factor on emission and adding it to fuel cost equation [2]. Various conventional techniques (e.g. Lambda iteration method) had been used to solve such problems but their performance and results were not up to the mark. Hence, many modern methods based on humans' evolution (e.g. Genetic Algorithm) and species' social behaviors (e.g. Particle Swarm Optimization) have been invented to cope with optimization problems. These methods have been very successful in terms of results and in terms of dealing with the complexities occurred in formulating such problems.

This work encompasses a brief introduction of different AI techniques popular to handle multi-objective optimization problems and summarize the results of various researchers who implemented CEED problem using these techniques. In addition to this, a self-implementation of CEED using two pioneer AI techniques; PSO and GA is done on IEEE 30-bus system with the inclusion of all emitted gases and the results are concluded.

## 2. PROBLEM FORMULATION

Combined economic emission dispatch in a thermal power plant is to minimize both fuel cost and pollutants emission simultaneously such that generation equals load demand with transmission losses and no unit violates its generation limits. Both fuel cost and pollutants emission can be equated as quadratic functions of generator real power. Our first objective is to minimize the cost function given as [1]:

$$F_T = \sum_{i=1}^{N} F_i(P_i) = \sum_{i=1}^{N}(a_i + b_i P_i + c_i P_i^2) \quad (1)$$

where $a_i$, $b_i$ and $c_i$ are the fuel cost coefficients of $i^{th}$ unit.

Our second objective is to minimize the emission function given as [1]:

$$E_T = \sum_{i=1}^{N} E_i(P_i) = \sum_{i=1}^{N}(\alpha_i + \beta_i P_i + \gamma_i P_i^2) \quad (2)$$

where $\alpha_i$, $\beta_i$ and $\gamma_i$ are the coefficients of generator emission characteristics of $i^{th}$ unit.

The minimization is bounded by following two constraints:

1. $$P_G = \sum_{i=1}^{N} P_i = P_D + P_L \quad (3)$$

where $P_G$ is total real power of $N$ generators, $P_D$ is total load demand and $P_L$ represents total transmission losses that can be calculated using B-coefficient matrix ($B_{mn}$) by the following relation:

$$P_L = \sum_{m=1}^{N}\sum_{n=1}^{N} P_m B_{mn} P_n \quad (4)$$

If losses are ignored, then

$$P_G = \sum_{i=1}^{N} P_i = P_D \quad (5)$$

2. All units should generate power within their minimum and maximum limits i.e.

$$P_{i,min} \leq P_i \leq P_{i,max} \qquad (6)$$

The objectives of minimizing fuel cost and emission can be made a single objective by the concept of penalty factor [1]:

$$\emptyset = k_1 F_T + k_2 h E_T \qquad (7)$$

where $F_T$ and $E_T$ are fuel cost and emission, $k_1$ and $k_2$ are constants having values 1 or 0 which will decide that either both dispatches are required or any one of them and $h$ is the penalty factor that is computed by first calculating $h_i$s by:

$$h_i = \frac{F_T(P_{i,max})}{E_T(P_{i,max})} = \frac{(a_i + b_i P_{i,max} + c_i P_{i,max}^2)}{(\alpha_i + \beta_i P_{i,max} + \gamma_i P_{i,max}^2)} \qquad (8)$$

These obtained $h_i$s are listed in increasing order and added with $P_{i,max}$ of every generator one by one starting from the first $h_i$ in the list until $\Sigma P_{i,max} \geq P_D$. When this condition becomes true, the reached $h_i$ in the list until that time is the price penalty factor $h$ for that particular load demand $P_D$ [2].

## 3. AI TECHNIQUES

A brief introduction of AI techniques reported in literature for multi-objective optimization problems in power systems is given below:

### i. Particle Swarm Optimization

In 1995, particle swarm optimization was invented by two scientists Kennedy and Eberhart. They were actually studying the patterns of social interaction within swarms of fishes and birds for searching food. Soon, they realized that by making certain changes, this social behaviour can be made a powerful optimizer [3].

In PSO, each particle has a position and its advancement in locality is decided by its velocity. Each particle is a valid solution to the problem and hence, its dimension is that of problem space. All the particles in a swarm modify their positions based on their individual best experiences and the best swarm experience in search of food or optimal solution in our case. The equations for velocity and position of a particle in a swarm are given in [3].

It has been observed that the simplest and efficient way of making classical PSO more effective is to use the constriction factor (CF) approach in which particle's velocity is multiplied by a parameter called constriction factor [4].

### ii. Genetic Algorithm [5]

D.E. Goldberg gave the basic theory for design and analysis of genetic algorithms based on the concept of biological evolution in 1988-89 and later, J.H. Holland established it systematically as a fact. In genetic algorithms, the problem variables are coded into binary strings. Each string is a valid solution to the problem and hence its length should be comparable to the problem space. Each individual (string) has a fitness value in the range 0-1 which basically relates that individual to the one having maximum fitness in that population. The initial population is randomly generated keeping in view the constraints but the next generations are produced by selection, crossover and mutation performed on the present one.

Selection is basically making a mating pool of fitter strings from present population. It is usually done by the concept of roulette-wheel. No new string is formed in selection phase. The greater the fitness of a string, the greater portion of the wheel's circumference it will occupy and the greater chance it will get to copy into mating pool. The wheel is spun $P$ times to select a population of good parents for producing off springs by crossover and mutation.

Crossover is performed on two parents of selected population to produce two off springs. Crossover site is selected randomly and the probability of crossover $(p_c)$ is usually taken higher. Finally, mutation is performed on the children produced after crossover which is just flipping of the child's bit at mutation site selected randomly. Its probability $(p_m)$ is usually taken lower. This journey of producing next generations continues until the optimum solution is found.

### iii. Artificial Bee Colony Optimization (ABC)

Dervis Karaboge modelled an optimization technique in 2005 naturally inspired from honeybees searching their food. The colony of artificial bees has three groups; employers, onlookers and scouts. Each group is in search of food but the basis of this search is different. Employed bees look for food sources depending on their individual experiences, onlooker bees do so by the collective experience of their hive mates and scouts choose food sources randomly out of any experience. Each food source is a solution to the problem and its nectar amount represents the quality of the solution. As each artificial bee is after a food source, hence the number of artificial bees in a colony equals number of food sources or possible solutions in the population. The employed bee whose food source is abandoned converts to a scout.

This algorithm starts with the random initialization of artificial bees. Then each onlooker bee selects a food source and keeps on modifying its position for the quality food search. If position cannot be improved further, it means that the food source is abandoned. In this case, scout bees discover new food sources and replace them with the abandoned ones. The respective equations are given in [6].

### iv. Ant Colony Search Algorithm (ACSA)

This algorithm was first proposed by Marco Dorigo in 1992 for his PhD thesis. His objective was to find the shortest path in a graph by modelling the behaviour of real ants in selecting a path between their colony and a source of food. Ants have the capability of following a shortest path without using visual cues. It is done by leaving pheromone trails like bread crumbs to signal directions to other ants. As a result, each ant probabilistically follows a direction rich in pheromones rather than a poorer one.

The algorithm starts with the initialization of ant number, states, iterations and parameters of the optimization problem under consideration. Let the ants finish their tour and their movement is governed by a probabilistic state transition rule that guides them to follow a path connected with shorter edges and rich in pheromones. Once all ants have completed their tour, their performance is tested by fitness function(s) of the optimization problem.

The next step is to update the pheromones of edges for each iteration. For the edges ants travelled, their pheromone intensity is updated by local and global updating rules. The local updating rule decides the patches within a path selected by an ant while the global updating rule governs best shorter tours among the ants on completion. The algorithm is converged and stopped when tour counter equals maximum iterations or all ants are making the same tours. The equations for state transition rule, local and global updating rules are stated in [7].

### v. Firefly Algorithm (FA)

This algorithm was first introduced by Xin-She Yang in 2009 inspired from social interaction of fireflies found in warm regions. It is a simple algorithm both from understanding and implementation point of view. The working of the algorithm depends upon the flashing characteristics of the lighting bugs. All fireflies have one gender and the attraction of one firefly towards the other is directly related to the brightness. Thus the firefly with less brightness moves towards the brighter one and it is true for any two fireflies. Their brightness increases as much as they come closer to each other. A firefly moves randomly if it do not find a brighter firefly in its surroundings. This brightness actually represents the fitness of the objective function to be optimized. The **attractiveness** among the fireflies is because of the brightness which in turn controls their **movement** and the **distance** among them. These three terms are mathematically described in [8].

### vi. Shuffled Frog Leaping Algorithm (SFLA)

This algorithm is inspired from evolution and social behaviour of species and hence a good blend of genetic algorithm and particle swarm optimization respectively. It consists of a set of interacting virtual frogs separated into different units. In each unit, simultaneously, a local search is carried out and then for global exploration of food place, the virtual frogs are periodically shuffled and rearranged into new units. As a result of this local and global search together, the probability of achieving the food place or best solution to the problem under consideration enhances greatly.

There are five steps in which this algorithm works:
1. Initialization
2. Fitness Evaluation
3. Partition of frogs into units.
4. Local search
5. Shuffling of frogs

The steps 3-5 keeps on iterating until the global evolutionary steps meet the maximum set limit [9].

### vii. Evolutionary Programming (EP)

EP is inspired by the biological evolution of human beings. It differs from the mainstream genetic algorithms by the fact that offsprings are created by mutation only and crossover is not considered. This makes it a leader among evolutionary computation methods in obtaining global optimal solution with lesser generations. The main steps are initialization, creation of offsprings by mutation and finally competition and selection of individuals to evaluate best solutions [10].

*Initialization:*
The initial population is randomly generated within allowed limits.

*Mutation:*
Offsprings are created by inverting one or more than one bits of parents' binary string. The proper selection of scaling factor has a huge influence on converging to global optimum. Small value of scaling factor results in premature convergence and its large value takes long execution time. Normally, constant scaling factor is used but step and nonlinear scaling factors are also reported in literature to improve the efficiency of EP algorithm.

*Competition and Selection:*
Parents and offsrings compete with each other for survival and individual with best fitness is selected as a parent for next generation. Initialization and mutation continue until stopping criteria is met.

### viii. Simulated Annealing (SA)

In 1983, Scott Kirkpatrick et al. described this algorithm as an optimization method. It is an adaptation of the Metropolis–Hastings algorithm and is inspired from a metallurgical process called annealing in which heat treatment and controlled cooling of solids is done to reduce their intrinsic defects by increasing their crystals size. It is an efficient method for problems that need a feasible good rather than best solution in a certain amount of time.

Initial temperature, iteration for each temperature, temperature decrement coefficient and maximum iterations hugely impact the efficiency of algorithm and the quality of the given solution [11]. Hence, these parameters should be decided and selected with deep study of the nature of the problem. The algorithm starts with generation of a random solution associated with initial temperature. The fitness of the solution is checked and the iteration process continues by discarding solutions with lesser quality compared to accepting fitter solutions until stopping criteria is met. In the start, the search space is wider and bad solutions are also welcomed. With the decrease of temperature, the algorithm narrows down its search towards zone with higher quality solutions and the acceptance for bad solutions drops down to zero.

### ix. Gravitational Search Algorithm (GSA)

This algorithm is a recent addition to the family of optimization algorithms by Rashedi et al. in 2009 based on laws of gravity and motion presented by Newton. In GSA, there are agents treated as objects and they attract each other due to gravity force and are pulled towards heavier objects. Each object is a solution or part of the solution to the problem and their mass is directly proportional to the quality of the solution. The algorithm starts with the initialization of positions of objects randomly in the search space. Then the positions of objects keep on changing according to their masses during each time cycle. This process continues until stopping criteria is met which is nothing but allotted time in this case. As soon as the allotted time finishes, algorithm stops and the latest solution is considered the optimal solution to the problem under consideration. The equations used in this algorithm are explained in [12].

### x. Biogeoraphy Based Optimization (BBO)

In 2008, Dan Simon proposed this optimization technique as a natural inspiration from the geographical division of organisms. In this algorithm, a geographical region for species is called habitat. A habitat containing large number of species is said to have High Suitability Index (HSI) and is calculated by independent variables or species of the habitat called Suitability Index Variables (SIVs). High HSI habitats have a low immigration rate but the emigration rate is higher due to congested space in the habitat. Hence, species tend to migrate from high HSI islands to low HSI islands and share their experiences with species of those islands. These rates of immigration and emigration represents number of species in the habitat.

The unique feature of this algorithm is that a good solution to a problem having high HSI once generated is not got destroyed. On the contrary, it shares its good features with solutions having low HSI and helps them to improve their quality. BBO basically is a blend of the good features of GA and PSO and it works based on migration and mutation described below [13]:

*Migration:*
In BBO, each solution or habitat can be modified based on certain probability. In order to retain the habitats having best HSI, the migration process from or to them is closed and it is open only for the islands having low HSI. The immigration rate of a habitat decides whether to modify it or not and the emigration rate of other habitats decides the island to which species from the habitat under modification have to migrate.

*Mutation:*
Each habitat has a probability representing quality of the solution. The mutation process gives chance of improvement to both solutions with high and low HSI. A check and balance on the habitats during the mutation process is necessary to keep the solutions with high HSI. Hence, if a habitat's HSI decreases after mutation, its previous version is restored to maintain the good features of high HSI habitat and later communicate them with those of low HSI until optimized solution is achieved.

## 4. REVIEW OF IMPLEMENTATION OF CEED USING AI TECHNIQUES

A summary of the efforts made by researchers in CEED case using different AI techniques is given below:

**i.　M. R. AlRashidi et al. [3]:**
They have done practical implementation of CEED with PSO successfully in 2006 considering almost all important constraints. It is a source reference for further research in this field after the historical published work on theoretical grounds by D. P. Kothari et al. They tested it on IEEE 30 bus system including all emitted gases and transmission line losses. They compared their results with a conventional method NR reported in literature and signified the effectiveness of AI techniques like PSO over conventional methods. Further, they opened door for advanced work on the handling of multi-objective function as a single objective by introducing the concept of weight factors which is refined as penalty factor approach in recent works.

**ii.　D. J. King et al. [14]:**
They implemented CEED on a 6-unit system with the inclusion of wind power modeled as a single generator. They discussed the issues that arise with this renewable energy integration and how they can be mitigated. GA and DSM were used and it was concluded that quality solution was obtained by GA but with slower computational time.

**iii.　G. P. Dixit et al. [6]:**
They implemented CEED considering $NO_X$ gas and line losses using ABC algorithm whose performance was checked on 3 and 6 generator system for different load demands. The results were compared with conventional method, SGA, RGA and Hybrid GA and they dictated the superiority of ABC over them with respect to global optimization, fast convergence and accuracy.

**iv.　R. Bharati et al. [7]:**
They tested CEED including transmission losses on IEEE 30 bus system with real coded GA, ACSA and conventional Lambda iteration method. Their simulation results indicated that GA's performance was in between Lambda iteration method and ACSA. The optimal solution was obtained by ACSA but with greater computation time compared to Lambda iteration method whose solution was the worst among the three methods used to solve this problem.
The effectiveness of this algorithm compared to PSO was also checked by M. S. Syed et al. [15] when they implemented CEED on 3 and 6 generator systems for different load demands. The results showed that ACSA is more efficient to give optimal operation cost for the combined dispatch.

**v.　O. Abedinia et al. [8]:**
They solved the CEED optimization problem using Firefly Algorithm for cost and emission dispatches separately and then selected the best compromised solution for combined dispatch. The results were compared with other modern techniques reported in literature (SPEA, NPGA, NSGA, MOPSO and MODE). IEEE 30 and 114 bus systems were considered and transmission losses were also incorporated. They concluded that Multi-Objective Firefly Algorithm (MOFA) gives the best compromised solution with respect to fuel cost, emission and line losses compared to other techniques.

**vi.　G. Chen [9]:**
He handled this multi objective problem as a single objective with the concept of price penalty factor. He tested the combined dispatch optimization problem on a 15-unit generator system with the help of GA, EP and SFLA and found that SFLA is dominant over others in terms of global optimization, computational accuracy and convergence rate.

**vii.　P. Venkatesh et al. [10]:**
They implemented separate and combined economic load/emission dispatches on IEEE 14, 30 and 118 bus systems using GA, micro GA and EP and found that EP was outstanding among them. Valve point loading and line flow constraints were also considered and the effectiveness of these modern techniques was checked by comparing their results with conventional approaches reported in the literature. A non-linear scaling factor was incorporated in EP and best optimal solution was given by EP as compared to GA, micro GA and other conventional techniques.

**viii.　P. Erdogmus et al. [11]:**
They compared the performance of GA and SA by applying them for separate and combined economic/environmental dispatches on IEEE 30 bus system for different load demands considering transmission losses. The results gave decision in favour of GA for separate dispatches but in case of combined dispatch, SA was superior to GA in terms of optimal solution.

**ix.　U. Guvenc et al. [12]:**
They used GSA to implement CEED on 4 benchmarked test systems (3, 10, 11 and 40 generator systems) including line losses in one and valve point loading in two of the systems. The simulation results of GSA were compared with γ-iteration, recursive, PSO, DE, GA, MODE, PDE, NSGA-II and SPEA 2 methods published in the literature. It was concurred that the proposed GSA for CEED is more effective and robust compared to other stochastic search algorithms in the literature.

**x. Bhattacharya et al. [16]:**
They proposed a variant of BBO called Oppositional BBO to implement CEED on a 3 generator system considering two gases and on a 6 generator system considering valve point loading. Transmission losses were included and the results of OBBO were compared with BBO, Tabu Search and NSGA-II in 3 generator system while with BBO and PSO in 6 generator system. In both cases, best solution was computed based on cost and emission independently and then a compromised best solution is chosen. They concluded that OBBO gives the optimal solution in less time compared to conventional BBO, Tabu Search, NSGA-II and PSO.

## 5. SIMULATION RESULTS

Combined economic emission dispatch using PSO and GA for 500 iterations has been implemented on MATLAB v8.2.0.701(R2013b) with Intel(R) Core(TM) i5-2410M CPU @ 2.30 GHz 2.30 GHz on 6 generators of IEEE 30 bus system for load demands of 1500 and 2000 MW. The initial parameters set in both algorithms are:

*PSO: Particles=10, $\mu_{max}$=0.9, $\mu_{min}$=0.4, $c_1$=2.05, $c_2$=2.05, $\varphi$=4.1, CF=0.7298*

*GA: Individuals=10, $p_c$=0.96, $p_m$=0.033*

The solutions with average operating time *(t)* have been selected out of 50 trials for comparison between PSO and GA.

The data for fuel cost and emission coefficients has been taken from [3]. Transmission line losses are not accounted for while all three gases ($NO_X$, $CO_X$ and $SO_X$) are considered; their penalty factors are calculated using (8) and the procedure following this equation, for load demands of 1500 and 2000 MW given as:

$P_D$=1500 MW: $h_{NOX}$=3.1669, $h_{COX}$=0.1221, $h_{SOX}$=0.9182
$P_D$=2000 MW: $h_{NOX}$=5.7107, $h_{COX}$=0.1307, $h_{SOX}$=0.9850

The results have been summarized in TABLE (all powers are in MW, emissions in Kg/h, costs in $/h and time in seconds) while the convergence characteristics of both algorithms for $P_D$ = 1500 MW and $P_D$ = 2000 MW have been shown in Fig. 1. and Fig. 2. respectively.

**TABLE**
**Simulation results of PSO and GA for IEEE 30 bus system with $P_D$ = 1500 MW and $P_D$ = 2000 MW**

| IEEE 30 | $P_D$ = 1500 MW | | $P_D$ = 2000 MW | |
|---|---|---|---|---|
| | PSO | GA | PSO | GA |
| P1 | 195.79 | 224 | 256.32 | 256 |
| P2 | 256.55 | 255 | 320.57 | 320 |
| P3 | 381.25 | 367 | 541.95 | 576 |
| P4 | 81.69 | 84 | 133.10 | 119 |
| P5 | 381.85 | 367 | 500 | 500 |
| P6 | 202.27 | 203 | 248.06 | 229 |
| $E_{NOX}$ | 1719.67 | 1748.28 | 2540.68 | 2573.48 |
| $E_{COX}$ | 39626.51 | 38949.38 | 73738.76 | 75848.25 |
| $E_{SOX}$ | 9623.04 | 9669.29 | 13601.05 | 13221.23 |
| EC | 19121.26 | 19171.65 | 37543.01 | 37631.82 |
| FC | 14827.57 | 14833.87 | 19445.29 | 19465.34 |
| **TC** | **33948.83** | **34005.52** | **56988.30** | **57097.16** |
| *t* | **0.3513** | **1.3030** | **0.3572** | **1.5351** |

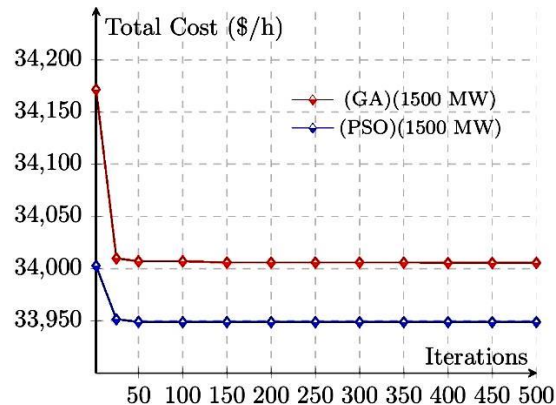

**Fig. 1. Convergence characteristics of PSO and GA for IEEE 30 bus system with $P_D$ = 1500 MW**

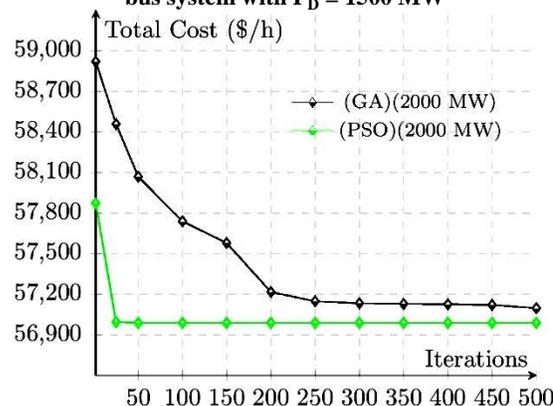

**Fig. 2. Convergence characteristics of PSO and GA for IEEE 30 bus system with $P_D$ = 2000 MW**

## 6. CONCLUSION

The simulation results dictates that PSO converges faster to global optimal solution as compared to GA due to its simple two equations (velocity and position) mathematical model while GA has to deal with three layers of operators (selection, crossover and mutation) in each iteration of the algorithm. These two methods are primary sources of many more AI techniques being reported in literature and a lot of research is going on to export their good qualities to make more variants and hybrid versions of AI techniques for increased performance.

Most of the fundamental AI techniques for complex optimization problems published in research are discussed in this work. Moreover, it is evident that these techniques can deliver an economical and safe solution for minimizing fuel cost and emission respectively on software level as compared to system modification such as Heat Recovery Steam Generator (HRSG) concept being practically deployed for the same purpose in combined cycle power plants.